\documentclass[runningheads]{llncs}
\usepackage{graphicx}
\usepackage{amssymb}
\usepackage{amsmath}

\usepackage{booktabs}
\usepackage{subcaption}
\usepackage{hyperref}
\usepackage[misc]{ifsym}

\usepackage{xspace,xcolor}
\newcommand{\name}{CASTNet\xspace}

\newcommand{\del}[1]{} 

\newcommand{\revmert}[1]{{#1}}

\makeatletter
\newcommand*\bigcdot{\mathpalette\bigcdot@{.6}}
\newcommand*\bigcdot@[2]{\mathbin{\vcenter{\hbox{\scalebox{#2}{$\m@th#1\bullet$}}}}}
\makeatother

\begin{document}
\title{CASTNet: Community-Attentive Spatio-Temporal Networks for Opioid Overdose Forecasting}
\titlerunning{CASTNet}
%
\author{Ali Mert Ertugrul\inst{1,2} \and
Yu-Ru Lin\inst{1}\textsuperscript{(\Letter)} \and
Tugba Taskaya-Temizel\inst{2}}
\authorrunning{A. M. Ertugrul et al.}
%
\institute{University of Pittsburgh, Pittsburgh PA, USA 
\email{\{ertugrul,yurulin\}@pitt.edu}
\and
Middle East Technical University, Ankara, Turkey
\email{ttemizel@metu.edu.tr}}
\toctitle{CASTNet: Community-Attentive Spatio-Temporal Networks for Opioid Overdose Forecasting}
\tocauthor{Ali Mert Ertugrul, Yu-Ru Lin, Tugba Taskaya-Temizel}

\maketitle

\begin{abstract}
Opioid overdose is a growing public health crisis in the United States. This crisis, recognized as ``opioid epidemic,'' has widespread societal consequences including the degradation of health, and the increase in crime rates and family problems. To improve the overdose surveillance and to identify the areas in need of prevention effort, in this work, we focus on forecasting opioid overdose using real-time crime dynamics. Previous work identified various types of links between opioid use and criminal activities, such as financial motives and common causes. Motivated by these observations, we propose a novel spatio-temporal predictive model for opioid overdose forecasting by leveraging the spatio-temporal patterns of crime incidents. Our proposed model incorporates multi-head attentional networks to learn different representation subspaces of features. Such deep learning architecture, called ``community-attentive'' networks, allows the prediction for a given location to be optimized by a mixture of groups (i.e., {\it communities}) of regions. In addition, our proposed model allows for interpreting what features, from what communities, have more contributions to predicting local incidents as well as how these communities are captured through forecasting. Our results on two real-world overdose datasets indicate that our model achieves superior forecasting performance and provides meaningful interpretations in terms of spatio-temporal relationships between the dynamics of crime and that of opioid overdose.

\end{abstract}
\section{Introduction}
\revmert{Opioid use disorders (OUD) and overdose rates in the United States have increased at an alarming rate since the past decade \cite{warner2011drug}. Overdose deaths have risen since the 1990s, and the number of heroin overdose deaths has risen sharply since 2010 \cite{rudd2016increases}. The growth rate of opioid overdose together with the number of impacted individuals in the U.S., has led many to classify this as an ``opioid epidemic" \cite{kolodny2015prescription}. Enhanced understanding of the dynamics of the overdose epidemic may help policy-makers to develop more effective epidemic prevention mechanisms and control strategies \cite{jalal2018changing}.}

\revmert{The opioid epidemic is a complex social phenomenon involving and interacting with various social, spatial and temporal factors \cite{burke2016forecasting}. Highlighting the links between opioid use and various factors has drawn significant attention. Studies have identified relationships between opioid use and crime incidences, including cause  \cite{bennett2008statistical}, effect \cite{hammersley1989relationship} and common causes \cite{seddon2005drugs}. Crime occurrences also have non-trivial spatio-temporal characteristics -- for example, routine activity theory suggested that crimes may exhibit spatio-temporal lags as the \textit{likely offenders} of one place may reach \textit{suitable targets} in other places. Therefore, how to unveil the complicated relationship between opioid use and crime incidences is challenging.
Moreover, detailed assessments of OUD and overdose growth require systematically collected well-resolved spatio-temporal data \cite{gruenewald2013b}. Yet, the amount of systematically monitored data either at a regional or local level in the U.S. is limited and there is no common reporting mechanism for incidents. On the other hand, crime data is meticulously collected and stored at finer-grained level. Given the plausible relationship between crime dynamics and opioid use as well as the availability of real-time crime data for various locations, in this study, we explore the capability of forecasting opioid overdose using real-time crime data.}

\revmert{Recent works in predictive modeling have shown significant improvement in spatio-temporal event forecasting and time series prediction \cite{qin2017dual,zhao2018distant}. However, these studies suffer from two main concerns. First, most of them overlook the complex interactions between local and global activities across time and space. Only a few have paid attention to this, yet they model the global activities as a single universal representation \cite{ertugrul2018forecasting,ertugrul2019}, which is either irrespective of event location or is reweighted based on a pre-defined fixed proximity matrix \cite{liang2018geoman}. None of them learns to differentiate the pairwise activity relationships between a particular event location and other locations. Second, most of the studies mainly focus on performance and lack interpretability to uncover the underlying spatio-temporal characteristics of the activities. Inspired by the idea of multi-head attentional networks \cite{vaswani2017}, we propose a novel deep learning architecture, called ``\name,'' for opioid overdose forecasting using spatio-temporal characteristics of crime incidents, which seeks to address the aforementioned problems. Assuming that different locations could share similar dynamics, our approach aims to learn different representation subspaces of cross-regional dynamics, where each subspace involves a set of locations called ``community'' sharing similar behaviors. The proposed architecture is ``community-attentive" as it allows the prediction for a given location to be individually optimized by the features contributed by a mixture of communities. Specifically, combining the features of the given target location and features from the communities (referred to as local and global dynamics), the model learns to forecast the number of opioid overdoses in the target location. Meanwhile, by leveraging a Lasso regularization \cite{scardapane2017group} and hierarchical attention mechanism, our method allows for interpreting what local and global features are more predictive, what communities contribute more to predicting incidences at a location, and what locations contribute more to each community.}

Overall, our contributions include: (1) \textit{A community-attentive spatio-temporal network:} We propose a multi-head attention based architecture that learns different representation subspaces of global dynamics (communities) to effectively forecast the opioid overdoses for different target locations.
(2) \textit{Interpretability in hierarchical attention and features:} First, \name incorporates a hierarchical attention mechanism which allows for interpreting community memberships (which locations form the communities), community contributions for forecasting local incidents and informative time steps in both local and global for the prediction. Second, \name incorporates Group Lasso (GL) \cite{scardapane2017group} to select informative features which succinctly captures what activity types at both local- and global-level are more associated with the future opioid overdoses. (3) \textit{Extensive experiments:} We performed extensive experiments using real-world datasets from City of Cincinnati and City of Chicago. The results indicate a significant improvement in forecasting performance with greater interpretability compared to several baselines and state-of-the-art methods.
\section{Related Work}
\revmert{The existing works have investigated the links between opioid use and various social phenomena as well as contextual factors including crime and economic stressors. Hammersley et al. \cite{hammersley1989relationship} stated that opportunities for drug use increase with involvement in criminal behavior. The people dependent on opiates are disproportionately involved in criminal activities \cite{bennett2008statistical} especially for the crimes committed for financial gain \cite{pierce2015quantifying}. Seddon et al. \cite{seddon2005drugs} revealed that crime and drug use share common set of causes and they co-occur together. Most of the works studying the relationship between opioid use and social phenomena employed basic statistical analysis, and focused on current situation and trends rather than predicting/forecasting overdose. Moreover, they overlooked the interactions among spatio-temporal dynamics of the locations. Among the studies predicting opioid overdose, \cite{kennedy2016opioid} have proposed a regression-based approach in state-level. Also, a neural network-based approach has been proposed \cite{ertugrul2018forecasting} to forecast heroin overdose from crime data, which identifies the predictive hot-spots. Yet, the effect of these hot-spots on prediction is universal and irrespective of event locations.}

\revmert{Furthermore, there have been studies that utilized spatial and temporal dependencies for event forecasting and time series prediction. Several studies employed neural models to forecast/detect events related to crime \cite{huang2018deepcrime} and social movements \cite{ertugrul2019}. Additionally, several studies utilized deep neural models for times series prediction. Ghaderi et al. \cite{ghaderi2017deep} proposed an RNN based model to forecast wind speeds. Qin et al. \cite{qin2017dual} presented a dual-stage attention-based RNN model to make time series prediction. Similarly, Liang et al. \cite{liang2018geoman} proposed multi-level attention networks for geo-sensory time series prediction. A few of the studies considered the complex relationships between local and global activities, yet they modeled the global activities as a universal representation, which either does not change from event location to location or is adjusted by a pre-defined fixed proximity matrix. Most of these works simply employed a single temporal model to model various local and global spatio-temporal activities, which is insufficient to capture the complex spatio-temporal patterns at both local and global levels. Moreover, existing methods primarily focus on forecasting performance, yet they provide no or limited interpretability capability to unveil the underlying spatio-temporal characteristics of the local and global activities.}
\section{Method}
\subsection{Problem Definition}
Suppose there are $L$ locations-of-interest (e.g. neighborhoods, districts) and each location $l$ can be represented as a collection of its static and dynamic features. While the static features (e.g. demographics, economical indicators) remain same or change slowly over a longer period of time, the dynamic features are the updates for each time interval $t$ (e.g. day, week). Let $X^{stat}_l$ be the static features of location $l$, and $X^{dyn}_{t, l}$ the set of dynamic features for location $l$ at time $t$. We are also given a discrete variable $y_{t^*, l} \in \mathbb{N}$ that indicates the number of opioid overdose incidents (e.g. emergency medical services (EMS) calls, deaths) at location $l$ at future time $t^*$. The collection of dynamic features from all locations-of-interest within an observing \textit{time window} with size $w$ up to time $t$ can be represented as $\mathcal{X}^{dyn}_{t-w+1:t} = \{\mathcal{X}^{dyn}_{t-w+1},\ldots,\mathcal{X}^{dyn}_t\}$, where $\mathcal{X}^{dyn}_{t'}=\{X^{dyn}_{t',1},\ldots, X^{dyn}_{t',L}\}$.

\revmert{Our goal is to forecast the number of opioid overdose incidents $y_{t^*,l}$ at specific location $l$ at a future time $t^*=t+\tau$, where $\tau$ is called the {\it lead time}. Forecasting is based on the static and dynamic features of the target location itself, as well as the dynamic features in the environment (from all locations-of-interest). Therefore, forecasting problem can be formulated as learning a function $f(X^{stat}_d, \mathcal{X}^{dyn}_{t-w+1:t}) \rightarrow y_{t^*,d}$ that maps the static and dynamic features to the number of opioid overdose incidents at future time $t^*$ at a {\it target} location $d$.}

To facilitate spatio-temporal interpretation of the forecasting, we seek to develop a model that can differentiate contribution of the features, the locality (local features vs. global features) and the importance of latent communities when contributing to the prediction of other locations. Therefore, we further organize the dynamic features $\mathcal{X}^{dyn}_{t-w+1:t}$ into two sets: the {\it local} features, $\{X^{dyn}_{t-w+1,d}, \ldots, X^{dyn}_{t,d}\}$ represent dynamic features for the target location $d$, and the {\it global} features, $\{X^{dyn}_{t-w+1,l}, \ldots, X^{dyn}_{t,l}\}$ for $l\in \{1, 2, \ldots, L\}$, contain the sequences of dynamic features for all locations of interest.
\begin{figure*}[t!]
\centering
\includegraphics[width=\linewidth]{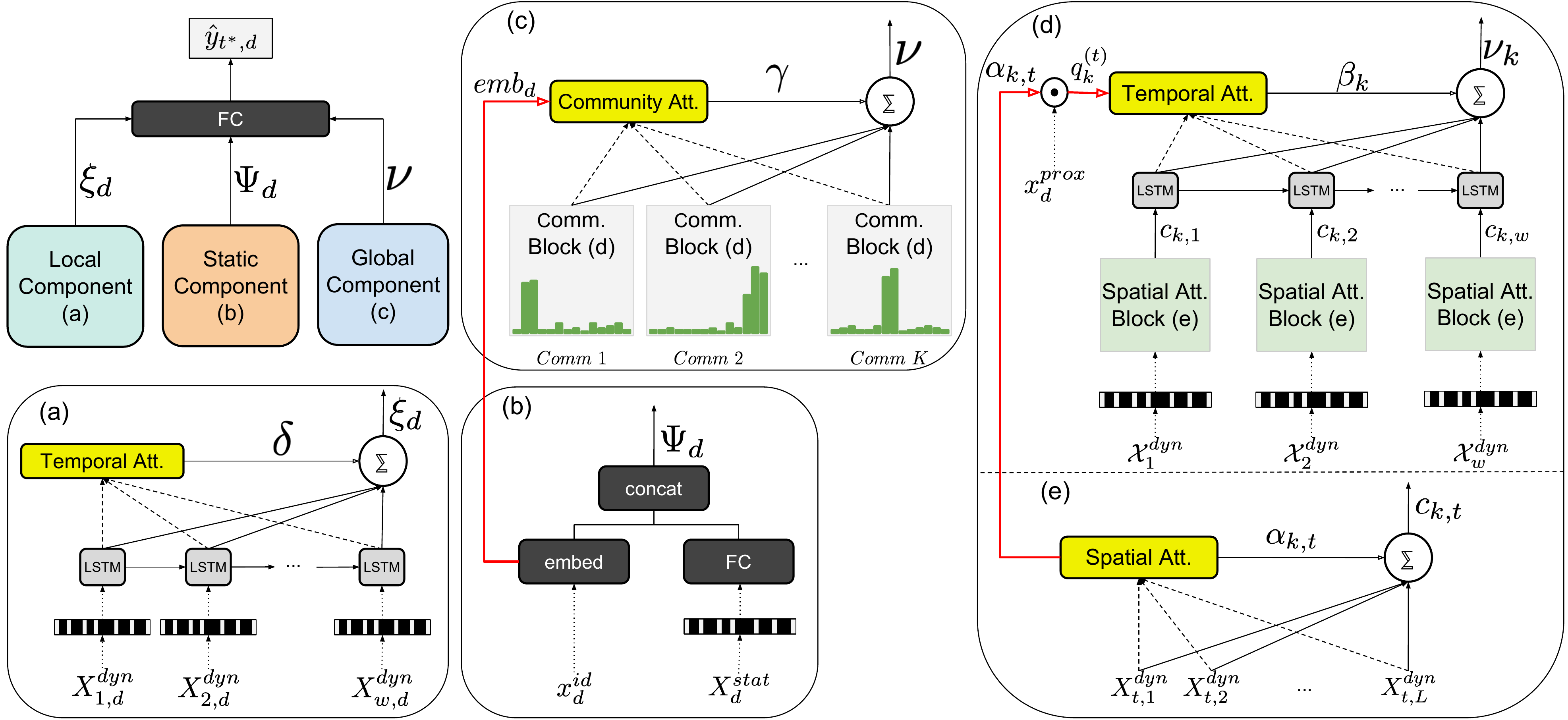}
\caption{
\label{fig:03_architecture}
\textbf{Overview of \name.} Local component (a) models local dynamics of the locations, and static component (b) models the static features. Global component (c) summarizes different representation subspaces (i.e. communities) of global dynamics, learned by community blocks (d), by querying these multi-subspace representations through the embedding of the target location ($emb_d$). Spatial Att. Block (e) reweights the global dynamics of locations. Checkered rectangles represent GL regularization. Red arrows indicate the queries for the corresponding attentions. ``FC": fully-connected layer; ``embed": embedding layer.}
\end{figure*}
\subsection{Architecture}
\revmert{We propose an interpretable, community-attentive, spatio-temporal predictive model, named \name. Our architecture consists of three primary components, namely local  (Fig.~\ref{fig:03_architecture}a), static  (Fig.~\ref{fig:03_architecture}b) and global  (Fig.~\ref{fig:03_architecture}c) components as follows:}
\subsubsection{Global Component.}
\label{sec:global_comp}
\revmert{This component produces the target location-specific global contribution to forecast the number of incidents at target location $d$ at future time $t^*$. It consists of $K$ number of community blocks where each community block learns a different representation subspace of the global dynamic features, which is inspired by the idea of multi-head attention \cite{vaswani2017}. A community block (Fig.~\ref{fig:03_architecture}d) models the global dynamic features through a hierarchical attention network which consists of a spatial attention block (Fig.~\ref{fig:03_architecture}e), a recurrent unit and a temporal attention. For clarity, we explain the internal mechanism of global component in a bottom-up manner in the order (Fig.~\ref{fig:03_architecture}e $\rightarrow$ \ref{fig:03_architecture}d $\rightarrow$ \ref{fig:03_architecture}c):}

\textit{\textbf{Spatial Attention Block}} is used to reweight the contribution of dynamic features of each location $i$ at time $t$. The attention weight, $\alpha_{k, t}^{(i)}$, represents the contribution of the location $i$ at time $t$ to the community $k$. Since higher spatial attention weight for a location indicates the involvement of its dynamic features in this community, we call this community membership. $c_{k, t}$ is the context vector, which summarizes the aggregated contribution of all locations as follows:
\begin{equation}
e_{k, t} = (v_k^{sp})^{\intercal} tanh(W^{sp}_k \mathcal{X}^{dyn}_t + b^{sp}_k)    
\end{equation}
\begin{equation}
\label{eq:_alpha}
\alpha_{k, t}^{(i)} = \dfrac{exp(e_{k, t}^{(i)})}{\sum_{l=1}^{L} exp(e_{k, t}^{(l)})}; ~~~
c_{k, t} = \sum_{l=1}^{L} \alpha_{k, t}^{(l)} X^{dyn}_{t, l}
\end{equation}
\noindent where $W^{sp}_k \in \mathbb{R}^{n \times n}$, $b^{sp}_k \in \mathbb{R}^{n}$ and $v_k^{sp} \in \mathbb{R}^{n}$ are the parameters to be learned, and $n$ is the dynamic feature size of any location. After the context vector $c_{k, t}$ is computed, it is fed to the recurrent unit.

\textit{\textbf{Recurrent unit}} is used to capture the temporal relationships among the reweighted global dynamic features for the community $k$ as $h_{k, t} = f_k(h_{k, t-1}, c_{k, t})$
where $f_k(.)$ is LSTM \cite{hochreiter1997} for community $k$, and $h_{k, t}$ is the $t$-th hidden state of $k$-th community. We use LSTM in our model (in each community block) since it addresses the vanishing and exploding gradient problems of basic RNNs.

\textit{\textbf{Temporal Attention}}
is applied on top of the LSTM
to differentiate the contribution of latent representations of global dynamic features at each time point and for each community. To make the output specific to target location, we incorporate a {\it query} scheme based on a time-dependent community membership (i.e., contribution of each location to the community) where the membership is further weighted based on the location's spatial proximity to target location (with nearby locations getting larger weights than the further ones). Specifically, let $\beta_{k}^{(i)}$ denotes the attention weight over the hidden state $h_{k, i}$ of community $k$ at time $i$. The context vector $\nu_{k}$, which is aggregate contribution from community $k$, can be learned through the proximity-based weighting scheme as:
\begin{equation}
q_{k}^{(i)} = x^{prox}_d \bigcdot \alpha_{k, i},
\end{equation}
\begin{equation}
\beta_{k}^{(i)} = \dfrac{exp(q_{k}^{(i)})}{\sum_{t=1}^{w} exp(q_{k}^{(t)})}, ~~~
\nu_{k} = \sum_{t=1}^{w} \beta_{k}^{(t)} h_{k, t},
\end{equation}

\noindent where $x^{prox}_d \in \mathbb{R}^{L}$ is a vector encoding the proximity of the target location $d$ to all locations. Here, the proximity of two locations is calculated based on the inverse of geographic distance (\textit{haversine}): \revmert{$prox(l_1, l_2) = \frac{1}{\sqrt{1 + dist(l_1, l_2)}}.$}

\textit{\textbf{Community Attention}}
aims to produce a global contribution with respect to the target location $d$ by combining different representation subspaces for each of the communities $\{\nu_1, \nu_2, \ldots, \nu_K\}$. A soft-attention approach is then employed to combine the contributions from all $K$ communities. Here, to make the prediction specific to the target location, we use a {\it query} scheme, which takes each community vector $\{\nu_k\}$ as a {\it key} and the target location embedding as a {\it query}:
\begin{equation}\label{eq:query}
u_{k} = r^{\intercal} tanh(V \nu_k + emb_{d}),
\end{equation}
\begin{equation}
\label{eq:_gamma}
\gamma^{(i)} = \dfrac{exp(u_{i})}{\sum_{k=1}^{K} exp(u_{k})}, ~~~
\nu = \sum_{k=1}^{K} \gamma^{(k)} \nu_{k},
\end{equation}
\noindent where $V \in \mathbb{R}^{m \times m}$, and $r \in \mathbb{R}^{m}$ are the parameters to be learned, $m$ is the number of hidden units in LSTMs, and $\nu$ is the output of the global component.

\subsubsection{Local Component.}
\label{sec:local_comp}
It is designed to model the contribution of local dynamic features for any target location $d$ (Fig.~\ref{fig:03_architecture}a). It includes a recurrent unit and a temporal attention that focuses on the most informative time instants. Dynamic features of target location are fed to the recurrent unit to model local dynamics as $s_{t} = g(s_{t-1}, X^{dyn}_{t, d})$
where $g(.)$ is LSTM, as in the global component, and $s_{t}$ is the $t$-th hidden state of LSTM. Then, we also employ a temporal attention on top of the LSTM in this component, which can select the most informative hidden states (time instants) with respect to the dynamic features of target location $d$. We only provide the calculation of output vector of the local component to be succinct as: $\xi_d = \sum_{t=1}^{w} \delta^{(t)} s_{t}$ where $\delta^{(t)}$ is the attention weight for the hidden state at time $t$, and $\xi_d$ is the output of the local component.

\subsubsection{Static Component.}
It models the static information specific to the target location (Fig.~\ref{fig:03_architecture}b). The input incorporates the static features, $X^{stat}_d$, and a one-hot encoding vector $x^{id}_d \in \mathbb{R}^{L}$ that represents the target location. We apply a fully connected layer (FC) to separately learn a latent representation for each of the two types of information. In particular, the one-hot location vector will be converted into an embedding $emb_d$ and will be utilized in the aforementioned query component (see Eq.~\eqref{eq:query}).
$\Psi_d$ is the output of this component, which is concatenation of learned embeddings and latent representation of static features.

\subsubsection{Objective Function.}
The objective function consists of three terms: prediction loss, orthogonality loss and Group Lasso (GL) regularization as follows:
\begin{equation}
\mathcal{L}_{overall} = \mathcal{L}_{predict} + \lambda \mathcal{L}_{ortho} + \eta \mathcal{L}_{GL},
\label{eq:loss}
\end{equation}
\noindent where $\lambda$ and $\eta$ are the tuning parameters for the penalty terms, and $\mathcal{L}_{predict} = \displaystyle\frac{1}{N}\sum_{i=1}^{N}(\hat{y}_i - y_i)^2$, is the mean squared error (MSE), $\hat{y}_i$ and $y_i$ are the predicted and actual number of opioid overdose incidents for sample $i$, respectively. A penalty term, $\mathcal{L}_{ortho}$ is added to avoid learning redundant memberships across communities, i.e., multiple communities may consist of a similar group of locations. To encourage community memberships to be distinguishable, we incorporate $\mathcal{L}_{ortho}$ into the objective function. Let $\bar{\alpha}_k$ be the community membership vector denoting how each location contributes to the community $k$, averaging over time, and $\Delta = \big[\bar{\alpha}_1, \bar{\alpha}_2,  \ldots, \bar{\alpha}_K\big] \in \mathbb{R}^{K \times L}$ is a matrix consisting of such membership vectors for all communities, the orthogonality loss is given by:
\begin{equation}
    \mathcal{L}_{ortho} = \Vert \Delta \cdot \Delta^{\intercal} - I \Vert_F^2,
\end{equation}
\noindent where $I \in \mathbb{R}^{K \times K}$ is the identity matrix. 
\revmert{This loss term encourages different communities to have non-identical locations as members as much as possible, which helps reduce the redundancy across communities. Lastly, we incorporate GL regularization into objective function, which imposes sparsity on a group level \cite{scardapane2017group}. Our main motivation to employ GL is to select community-level and local-level informative features. It enables us to interpret and differentiate which features are important for opioid overdose incidents. It is defined as:}
\begin{equation}
    \mathcal{L}_{GL} = \sum_{k=1}^{K} \left(\Vert Z_k^{glob} \Vert_{2,1}\right) + \Vert Z^{local} \Vert_{2,1} + \Vert Z^{stat} \Vert_{2,1},
\end{equation}
\begin{equation}
    \Vert{Z}\Vert{}_{2,1} = \sum_{g \in G} \sqrt{\mid g \mid} \Vert g \Vert_{2},
\end{equation}
\noindent where $Z_k^{glob}$ denotes input weight matrix in the $k^{th}$ community block in global component. $Z^{local}$ and $Z^{stat}$ are input weight matrices in the local and static components, respectively. $g$ is vector of outgoing connections (weights) from an input neuron, $G$ denotes a set of input neurons, and ${|g|}$ is the dimension of $g$.
\subsection{Features}
\revmert{\noindent \textbf{Static features} are 9 features from the census data related to economical status (median household income, per capita income, poverty), housing status (housing occupancy and housing tenure), educational level (\% of high school graduation and below) and demographics (population, gender and race diversity index).}

\revmert{\noindent \textbf{Dynamic features} are to capture the crime dynamics of the locations that may be predictive for opioid overdose. We extract them from public safety data portals of the cities. The crime data gathered from different cities may have different categories. We consider the highest level, ``primary crime types'' and eliminate rare ones. Crime categories used in this work can be found in Fig.~\ref{fig:05_dynamic_features}. In addition to total number of incidents for each primary crime type, we also use total number of crime and opioid overdose incidents as dynamic features. We extract 14 and 9 crime-related dynamic features for Chicago, and Cincinnati, respectively. Together with the number of opioid overdose incidents, the total number of dynamic features are 15 for Chicago and 10 for Cincinnati.}
\section{Experiments}
\label{sec:exp}
\subsection{Datasets}
\revmert{We apply our method on two cities, Chicago and Cincinnati. We used ``Statistical Neighborhood Approximations'' of Cincinnati and ``community areas'' of Chicago as ``neighborhoods''. There are 77 and 50 neighborhoods in Chicago and Cincinnati, respectively. While we select 47 neighborhoods from Chicago (where $\sim80\%$ of opioid overdose deaths occur), we use all neighborhoods of Cincinnati. Chicago dataset spans (08/03/15 - 08/26/18) and contains 573207 crimes and 1468 opioid overdose deaths. Cincinnati dataset spans (08/01/15 - 06/01/18) and contains 75779 crimes and 5401 EMS responses. We collect the following data:}

\revmert{\noindent \textbf{Crime data:} We collect crime incident information (geo-location, time and primary type of the crimes) from the open data portals of the cities. We use Public Safety Crime dataset\footnote{https://data.cityofchicago.org/Public-Safety/Crimes-2001-to-present/ijzp-q8t2} and  Police Data Initiative (PDI) Crime Incidents dataset\footnote{https://data.cincinnati-oh.gov/Safer-Streets/PDI-Police-Data-Initiative-Crime-Incidents/k59e-2pvf} to extract such information for Chicago and Cincinnati, respectively.}

\noindent \textbf{Opioid overdose data:} We collect different types of opioid overdose data for each city since there is no systematic monitoring of drug abuse at either a regional or state level in the U.S. For Chicago, we collect opioid overdose death records (geo-location and time) from Opioid Mapping Initiative Open Datasets\footnote{https://opioidmappinginitiative-opioidepidemic.opendata.arcgis.com/}. On the other hand, we utilize the EMS response data\footnote{https://insights.cincinnati-oh.gov/stories/s/Heroin/dm3s-ep3u/} for heroin overdoses in Cincinnati. 

\revmert{\noindent \textbf{Census data:} We use 2010 Census data to extract features about demographics, economical status, housing status and educational status of the neighborhoods.}

\subsection{Baselines}
\label{sec:_baselines}
\revmert{We compare our model with a number of baselines as follows: \textbf{HA}: Historical average, \textbf{ARIMA}: a well-known method for predicting future values for time series, \textbf{VAR}: a method that captures the linear inter-dependencies among multiple time series and forecasts future values, \textbf{SVR}: two variants of Support Vector Regression; SVR$_{ind}$ (trained separate models for each location) and SVR$_{all}$ (trained a single model for all locations), \textbf{LSTM}: a network in which dynamic features are fed to LSTM, then the latent representations are concatenated with static features for prediction, \textbf{DA-RNN} \cite{qin2017dual}: a dual-staged attention-based RNN model for spatio-temporal time series prediction, \textbf{GeoMAN} \cite{liang2018geoman}: a multi-level attention-based RNN model for spatio-temporal prediction, which shows state-of-the-art performance in the air quality prediction task, \textbf{ActAttn} \cite{ertugrul2019}: a hierarchical spatio-temporal predictive framework for social movements.

Furthermore, to evaluate the effectiveness of individual components of our model, we also include its several variants for the comparison: \textbf{\name-noGL}: GL regularization is not included in the loss function, \textbf{\name-noOrtho}: Orthogonality penalty is not applied so that differentiation of the communities is not encouraged, \textbf{\name-noSA}: Spatial attentions are removed from the community blocks. Instead, the feature vectors of all locations are concatenated, \textbf{\name-noTA}: The temporal attentions in both local and global components are removed from the architecture, \textbf{\name-noCA}: Community attention is removed from the architecture. Instead, the context vectors of the communities are concatenated. \textbf{\name-noSC}: The static features are excluded from the architecture, yet the location-ID is still embedded.}

\revmert{\noindent \textbf{Settings:}}
We used `week' as time unit and `neighborhood' as location unit. We divided datasets into training, validation and test sets with ratio of 75\%, 10\% and 15\%, respectively. We set $\tau = 1$ to make short-term predictions. For RNN-based methods, hidden unit size of LSTMs was selected from $\{8, 16, 32, 64\}$. The networks were trained using Adam optimizer with a learning rate of 0.001. For each LSTM layer, dropout of 0.1 was applied to prevent overfitting. In our models, the regularization factors $\lambda$ and $\eta$ were optimized from the small sets $\{0.001, 0.005, \ldots, 0.05\}$ and $\{0.001, 0.0015, \ldots, 0.01\}$, respectively using grid search. For \textit{ARIMA} and \textit{VAR}, the orders of the autoregressive and moving average components were optimized for the time lags between 1 and 11. For RNN-based methods, we performed experiments with different window sizes $w \in \{5, 10, 15, 20\}$, and shared the results for $w=10$ (the best setting for all models). Our code and data are available at \url{https://github.com/picsolab/castnet}.
\section{Results}
\begin{table}[t!]
\centering
\caption{Performance Results.}
\label{tab:perf_results}
\begin{tabular}{@{}lcccccccc@{}}
\toprule
 &  & \multicolumn{3}{c}{Chicago} &  & \multicolumn{3}{c}{Cincinnati} \\ \cmidrule{3-5} \cmidrule{7-9}
 &  & \multicolumn{1}{c}{MAE} &  & \multicolumn{1}{c}{RMSE} & $\quad$ & MAE &  & RMSE \\ \cmidrule{3-3} \cmidrule{5-5}  \cmidrule{7-7} \cmidrule{9-9}
HA &  & 0.2329 & & 0.3385 &  & \multicolumn{1}{c}{0.5728} & & \multicolumn{1}{c}{0.8727} \\
ARIMA &  & 0.2272 & & 0.3396 &  & \multicolumn{1}{c}{0.5717} & & \multicolumn{1}{c}{0.8952} \\
VAR & & 0.2242 & & 0.3386 & & \multicolumn{1}{c}{0.5606} & & \multicolumn{1}{c}{0.8712} \\
SVR$_{ind}$ & & 0.2112 & & 0.3321 & & 0.5153 & & 0.8609 \\
SVR$_{all}$ & & 0.1984 & & 0.3063 & & 0.4886 & & 0.8602 \\
LSTM & & 0.2024 & & 0.3134 & & 0.5235 & & 0.8267 \\
DA-RNN \cite{qin2017dual} & & 0.1726 & & 0.3051 & & 0.4817 & & 0.8225 \\
GeoMAN \cite{liang2018geoman} & & 0.1679 & & 0.2829 & & 0.5034 & & 0.8453 \\
ActAttn \cite{ertugrul2019} & & 0.1693 & & 0.2937 & & 0.4827 &  & 0.8326 \\
\midrule
\name-noGL & &  0.1662 & & 0.3129 & & 0.4703 & & 0.8311 \\
\name-noOrtho & & 0.1649 & & 0.2948 & & 0.4716 & & 0.8109 \\
\name-noSA & & 0.1608 & & 0.2893 & & 0.4579  & & 0.8152 \\
\name-noTA & & 0.1641 &  & 0.2876 & & 0.4700 & & 0.8141 \\
\name-noCA & & 0.1631 & & 0.3069 & & 0.4730 & & 0.8225 \\
\name-noSC & & 0.1693 & & 0.2980 & & 0.4692 & & 0.8291 \\
\midrule
\name & & \textbf{0.1391} & & \textbf{0.2679} & & \textbf{0.4516} & & \textbf{0.8032} \\
\bottomrule
\end{tabular}
\end{table}
\subsection{Performance Comparison}
Table \ref{tab:perf_results} shows that \name achieves the best performance in terms of both mean absolute error (MAE) and root mean squared error (RMSE) on both datasets. Our model shows 17.2\% and 5.3\% improvement in terms of MAE and RMSE, respectively, on Chicago dataset compared to state-of-the-art approach GeoMAN. Similarly, \name enhances the performance 6.3\% and 2.4\% on Cincinnati dataset in terms of MAE and RMSE, respectively, compared to DA-RNN. Furthermore, we observe that mostly spatio-temporal RNN-based models outperform other baselines, which indicates they better learn the complex spatio-temporal relationships between crime and opioid overdose dynamics.

\revmert{We evaluate the effectiveness of each individual component of \name with an ablation study. As described in Section~\ref{sec:_baselines}, each variant is different from the proposed \name by removing one tested component. Table \ref{tab:perf_results} shows that the removal of GL results in a significantly lower performance compared to the others. In addition, \name-noGL cannot select informative features. Excluding orthogonality term (\name-noOrtho) loses the ability to learn distinguishable communities and reduces the performances. Comparing \name with \name-noCA shows the impact of community attention on the performance, indicating that learning pairwise activity relationships between an event location and communities is crucial. Location-specific static features are informative since their exclusion (\name-noSC) degrades the performance in both cases. Spatial attention provides the least performance gain for both cases, yet, its removal (\name-noSA) results in loss of interpretability. These results reflect that each individual component significantly contributes to the performance.}

\begin{figure}[t!]
    \centering
    \begin{subfigure}[t]{0.4\textwidth}
        \includegraphics[clip, trim=0.5cm 0.5cm 0.5cm 0.5cm, width=1.00\textwidth]{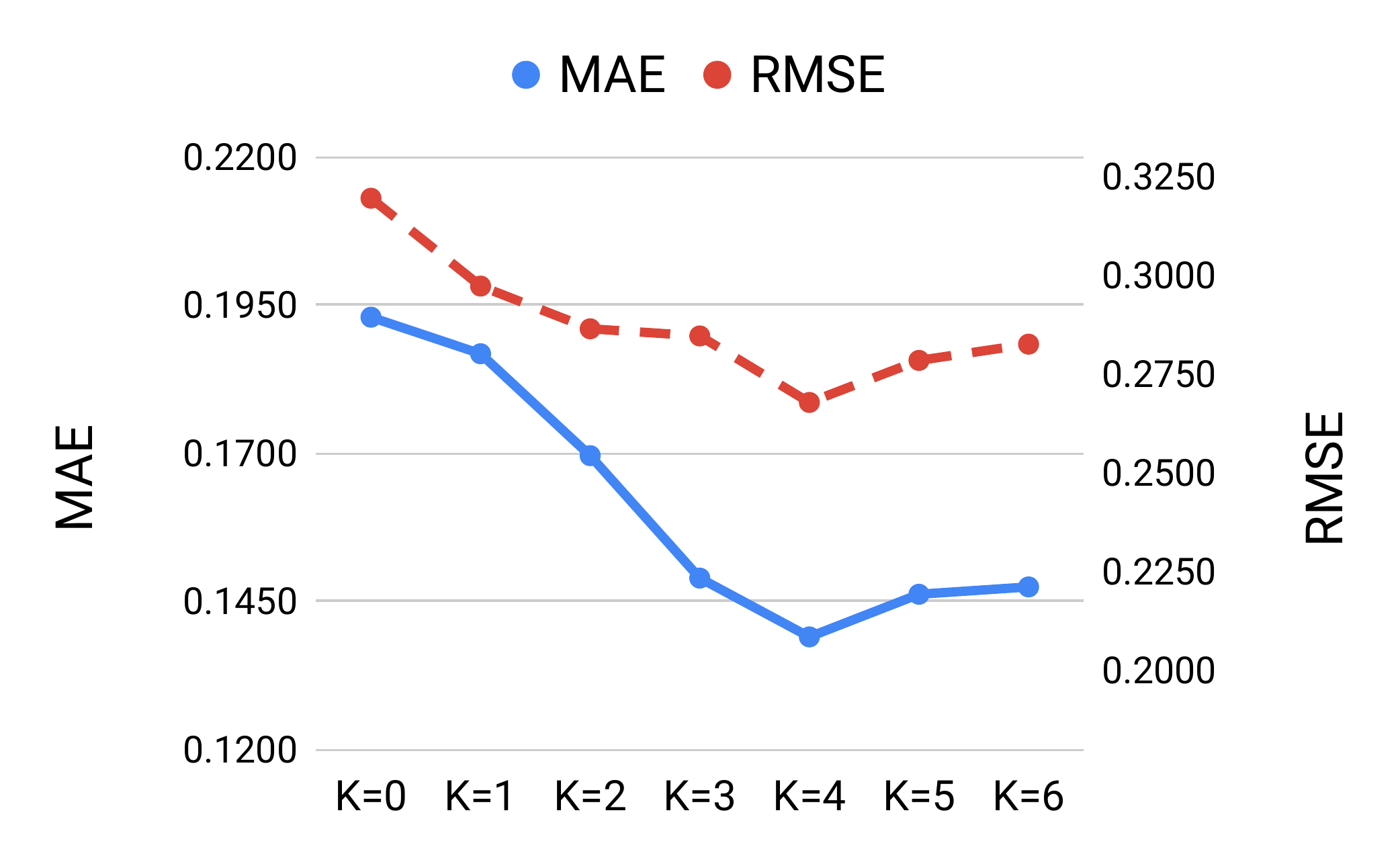}
        \caption{Chicago}
        \label{fig:05_chi_k_analysis}
    \end{subfigure}
    \begin{subfigure}[t]{0.4\textwidth}
        \includegraphics[clip, trim=0.5cm 0.5cm 0.5cm 0.5cm, width=1.00\textwidth]{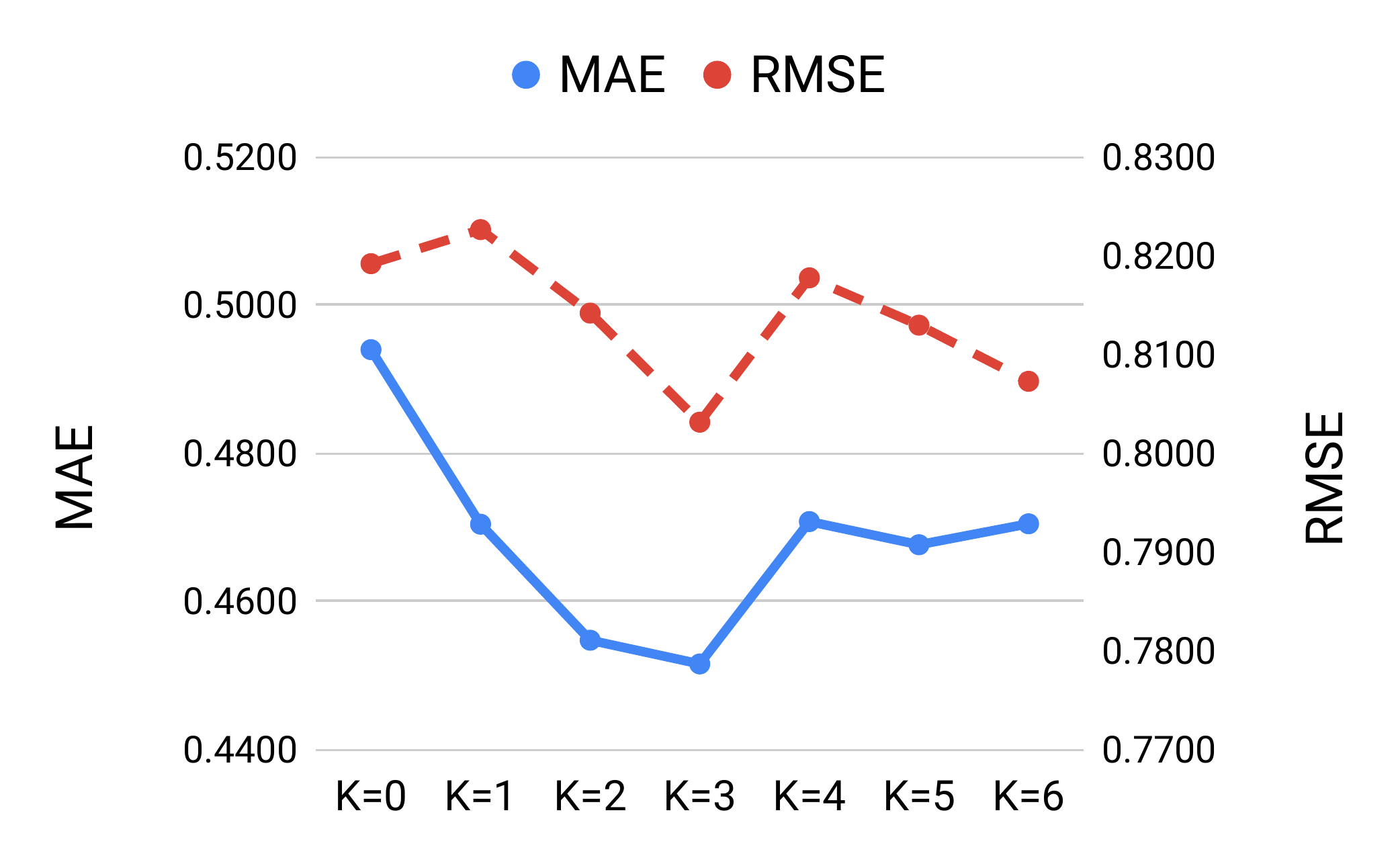}
        \caption{Cincinnati}
        \label{fig:05_cin_k_analysis}
    \end{subfigure}
    \caption{MAE and RMSE results w.r.t change in the number of communities $K$.}
    \label{fig:05_k_analysis}
\end{figure}

\begin{figure}[t!]
    \centering
    \begin{subfigure}[t]{0.375\textwidth}
        \includegraphics[clip, trim=0.1cm 0.6cm 0.1cm 0.1cm, width=.9\textwidth]{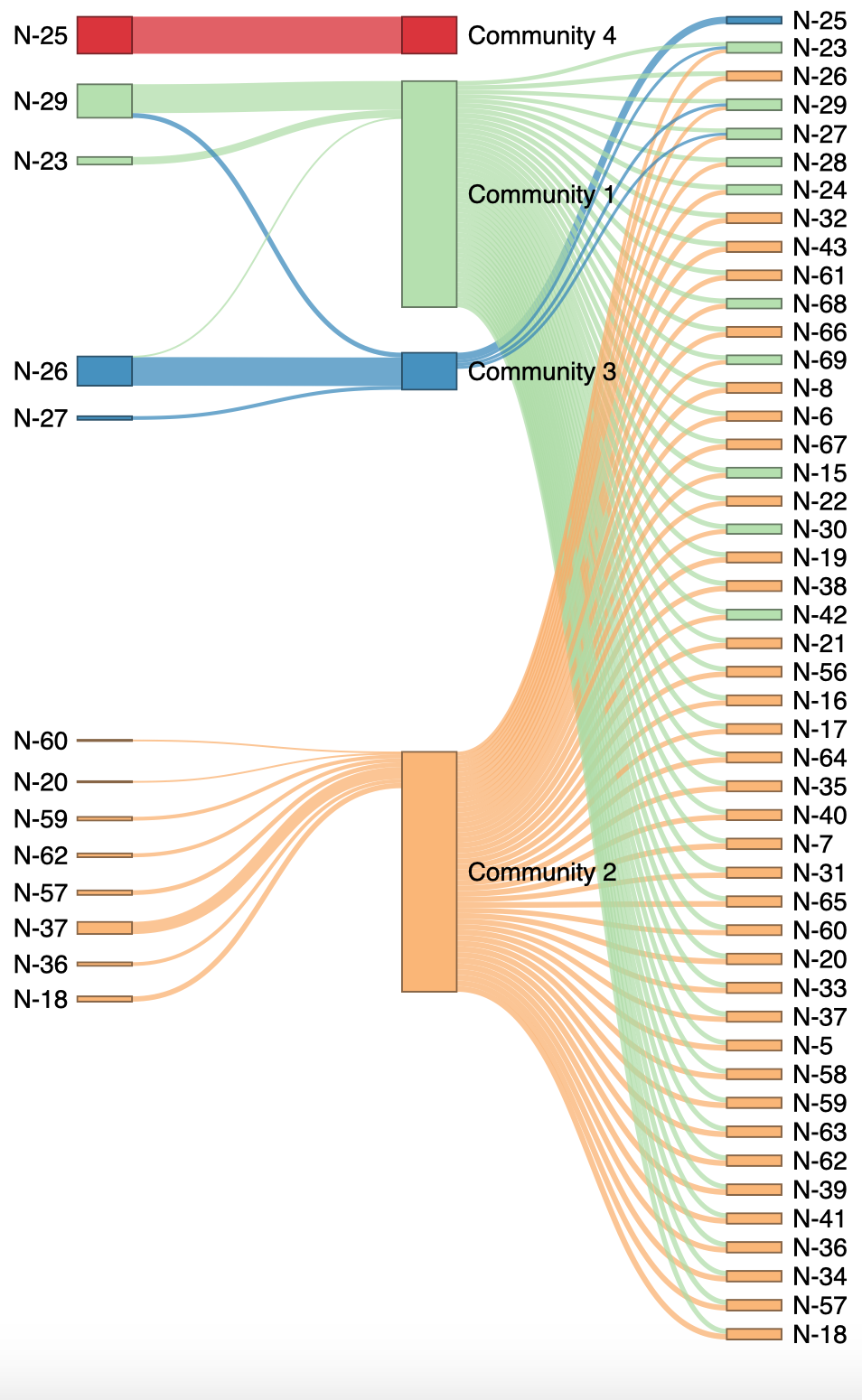}
        \caption{Chicago}
        \label{fig:05_chi_sankey}
    \end{subfigure}
    \begin{subfigure}[t]{0.38\textwidth}
        \includegraphics[clip, trim=0.1cm 0.6cm 0.1cm 0.1cm, width=.9\textwidth]{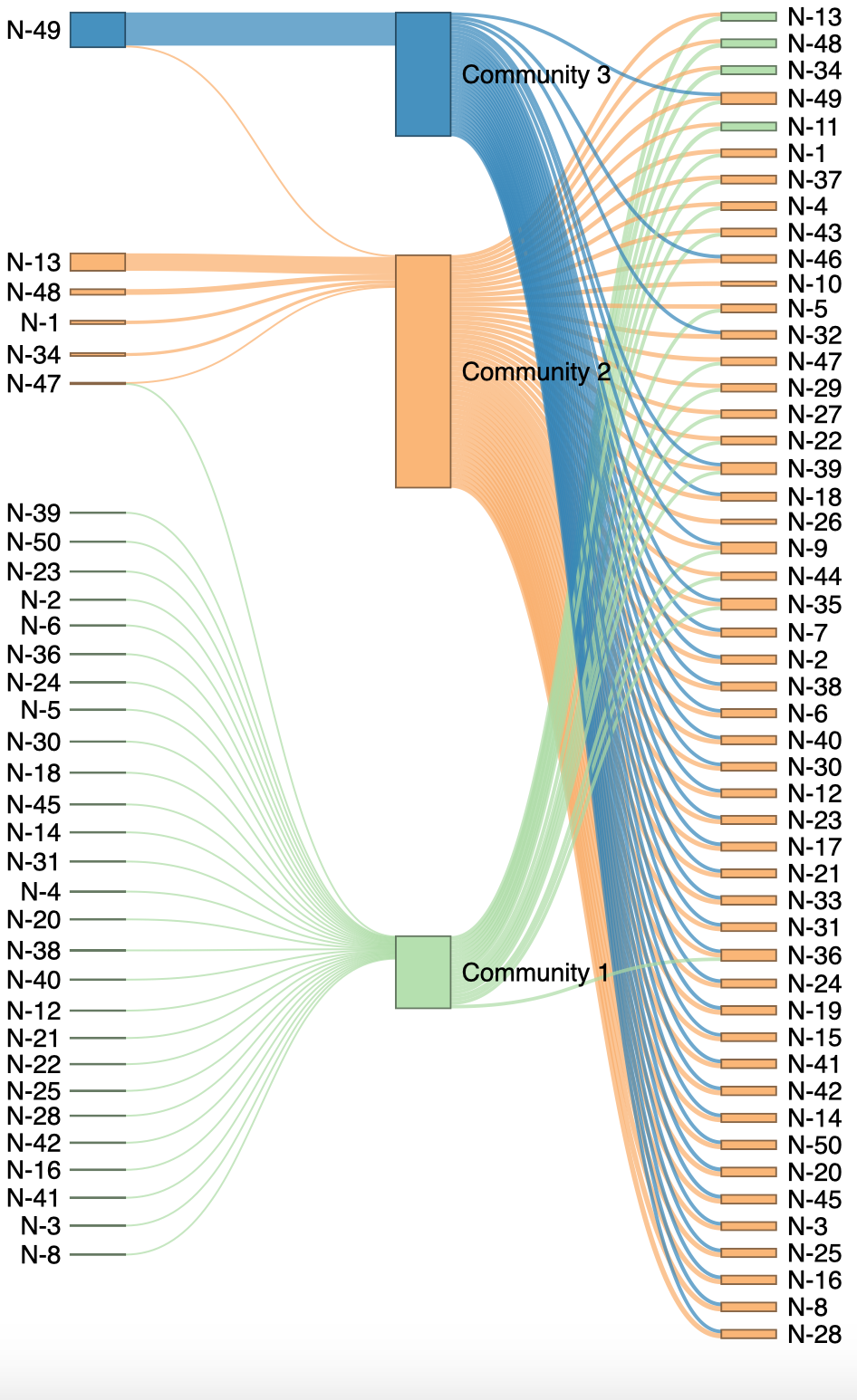}
        \caption{Cincinnati}
        \label{fig:05_cin_sankey}
    \end{subfigure} 
    \caption{\textbf{Community memberships and community contributions on forecasting.} 
    For each community, left side represents community memberships (how each location contributes to the community), and right side represents the average community contribution (how the community contribute to predicting a target location). Edge thickness indicates the weight of community membership (left side) and community contribution (right side). Node size denotes overall community membership of a location (left side) and overall community contribution to forecasting overdose (right side) in the target neighborhood. Edge color shows the input and output of a specific community. Node color of a neighborhood indicates the community for which the corresponding neighborhood has the highest membership (left side). Node color of a neighborhood denotes the community from which the neighborhood takes the largest contribution (right side). Edges whose weights are above a certain threshold are shown.}
    \label{fig:05_sankey}
\end{figure}

\revmert{We further evaluate the performance of \name with respect to the change in number of communities $K$. We report results for $K\in\{0, 1, \ldots, 6\}$ in Fig.~\ref{fig:05_k_analysis}. When $K=0$, the model ignores global contribution, and when $K=1$, the model yields a single universal representation of global contributions, which is irrespective of event locations. The best performances are obtained when $K=4$ for Chicago and $K=3$ for Cincinnati. As $K$ increases until the optimum value, the performance increases, and some communities are decomposed to form new ones. After the optimum value of $K$, performance starts to decrease slightly or remains stable, and the semantic subspaces of some communities become similar. With this experiment, we indicate that learning different representations of global activities significantly improves the forecasting performance.}
\revmert{\subsection{Community Memberships and Community Contributions}
We analyze community memberships of the neighborhoods and community contributions on forecasting opioid overdose by answering the following questions.}
\revmert{\noindent \textbf{How do locations contribute to communities?}}
\name learns different representation subspaces (communities) of global dynamics unlike the previous work \cite{liang2018geoman,ertugrul2019}, and each community consists of a group of different members due to orthogonality penalty. 
\revmert{We represent the learned communities and their memberships (i.e., the spatial attention weights $\alpha$ in Eq.~\eqref{eq:_alpha}, averaged over time for ease of interpretation) on the left sides of Fig.~\ref{fig:05_chi_sankey} and \ref{fig:05_cin_sankey} for Chicago and Cincinnati, respectively. Neighborhoods on the left sides of Fig. \ref{fig:05_chi_sankey} and Fig. \ref{fig:05_cin_sankey} are ordered by the number of crimes. As shown in Fig.~\ref{fig:05_sankey}, most locations have dedicated to one community. For Chicago model 
(Fig.~\ref{fig:05_chi_sankey}), \textit{Austin (25)}, which has the highest number of crime incidents and opioid overdose deaths, formed a separate community $C_4$ by itself. While \textit{North Lawndale (29) and Humboldt Park (23)} together formed the community $C_1$, \textit{West Garfield Park (26)}, \textit{East Garfield Park (27)} and \textit{North Lawndale (29)} formed $C_3$. Note that neighborhoods of $C_1$ and $C_3$ have the highest opioid overdose death rate after \textit{Austin (25)}. On the other hand, $C_2$ is formed by the neighborhoods having low crime and overdose death rates including \textit{Fuller Park (37), McKinley Park (59)} and \textit{West Elsdon (62)}. Furthermore, for Cincinnati model (Fig.~\ref{fig:05_cin_sankey}), \textit{Westwood (49)}, where the highest number of crimes were committed, formed a separate community $C_3$ by itself similar to \textit{Austin (25)} in Chicago. \textit{East Price Hill (13), West Price Hill (48), Avondale (1)} and \textit{Over-The-Rhine (34)} formed $C_2$ where these neighborhoods have the highest crime rate after \textit{Westwood (49)} and the highest opioid overdose rate. $C_1$ is formed by rest of the neighborhoods (with low and moderate crime rates) and their memberships of that community are almost equal.}

\revmert{\noindent \textbf{How do the communities contribute to forecasting?}}
\revmert{\name is capable of modeling the pairwise activity relationships between a particular event location and the communities. It allows the target location to attend the communities to select location-specific global contributions. We analyze how these communities contribute to forecasting by visualizing the community attention weights ($\gamma$ in Eq.~\eqref{eq:_gamma} averaged over test samples for each neighborhood) in Fig.~\ref{fig:05_chi_sankey} and Fig.~\ref{fig:05_cin_sankey} for Chicago and Cincinnati, respectively. The right sides of the figures indicate the average community contributions for the neighborhoods, which are ordered by the number of opioid overdoses on the right sides. For Chicago, $C_1$ and $C_2$ have more contributions than others on forecasting overdose. While $C_2$ contributes more to neighborhoods with low or moderate opioid overdose death rate, $C_1$ and $C_3$ contribute more to the neighborhoods with higher death rate meaning that any neighborhood attends more to the community, which is formed by the similar neighborhoods. $C_4$ does not significantly contribute to any neighborhood although it is formed by a crime hot-spot (\textit{Austin (25)}). For Cincinnati, $C_2$ is very dominant and makes the largest global contribution to most of the neighborhoods. The neighborhoods that formed $C_2$ and $C_3$ (e.g. \textit{East Price Hill (13), West Price Hill (48), Westwood (49)}) are very predictive, and the change in their dynamics have greater impact on forecasting overdoses in the target neighborhoods. On the other hand, $C_1$ has larger contribution to neighborhoods with the highest overdose rate indicating that crimes committed in the members of $C_1$ are informative for forecasting overdoses in opioid hot-spots.}
\subsection{Feature Analysis}

\begin{figure*}[t!]
\centering
\includegraphics[clip, trim=0cm 1.25cm 3cm 2.5cm, width=.8\textwidth]{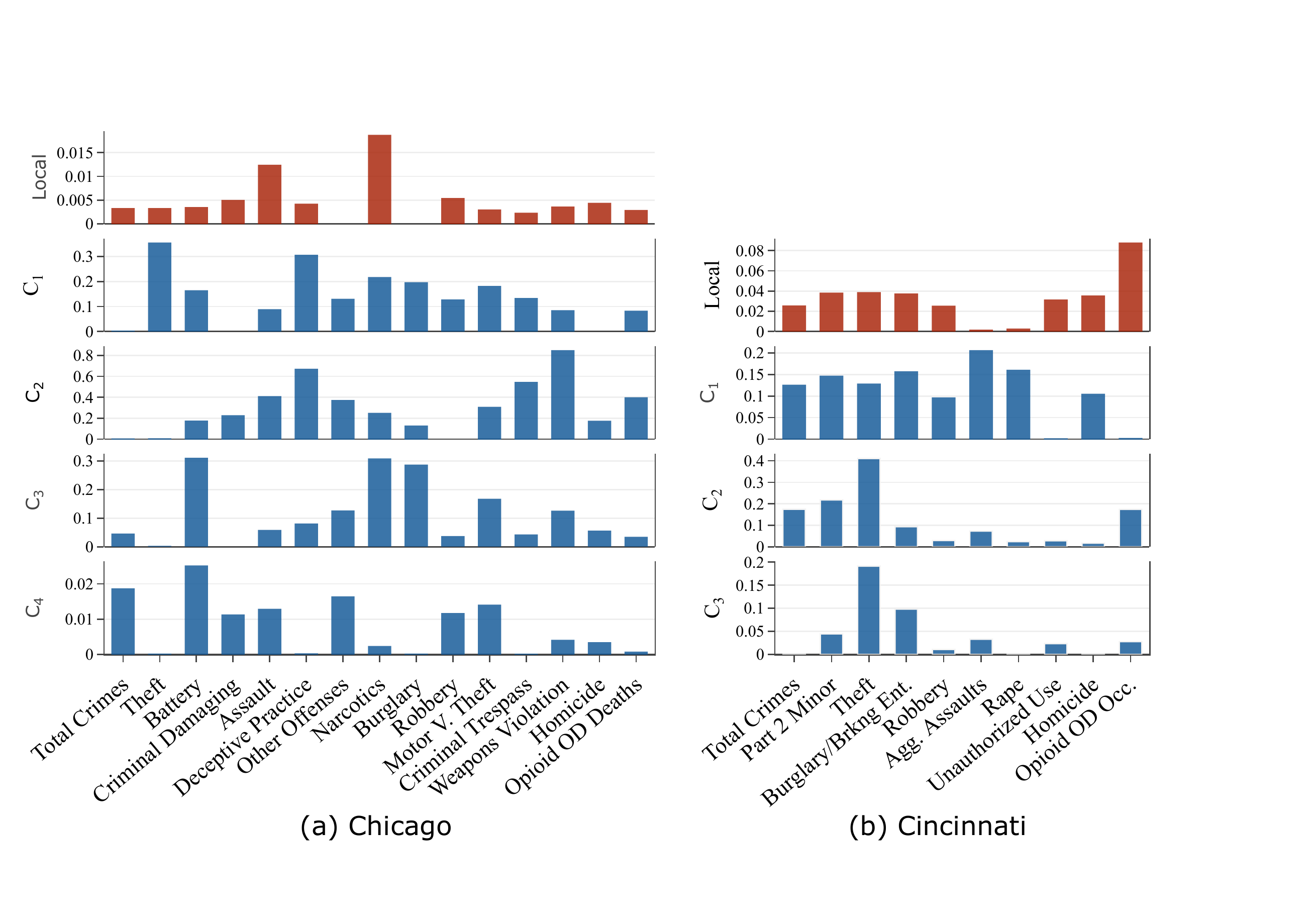}
\caption{\label{fig:05_dynamic_features}
\textbf{Importance of dynamic features.} Mean absolute values of input weights of local and global components.}
\end{figure*}

We investigate the importance of \textbf{dynamic features} by analyzing the mean absolute input weights of local and global components as shown in Fig.~\ref{fig:05_dynamic_features}. For Chicago case, GL selects \textit{Narcotics} and \textit{Assault} as the most important features for future opioid overdose deaths in the same location. Moreover, \textit{Theft, Deceptive Practice, Narcotics, Burglary} and \textit{Motor V. Theft} are the predictive features from $C_1$ while \textit{Weapons Violation, Deceptive Practice (e.g. Fraud)} and \textit{Criminal Trespass} are significant from $C_2$. Recall that, $C_1$ and $C_2$ are the most contributing communities to forecasting (see Fig.~\ref{fig:05_chi_sankey}). This shows that property crimes (\textit{e.g. Theft, Burglary, Deceptive Practice}) are more significant predictors than the violent crimes for Chicago. Such crimes previously committed in the members of $C_1$ and $C_2$ may be a significant indicator of future opioid overdose deaths in Chicago. On the other hand, \textit{Battery, Narcotics, Burglary, and Motor V. Theft} are predictive features from $C_3$ while \textit{Battery, Total Crimes} and \textit{Other Offenses (e.g. offenses against family)} are significant from $C_4$. However, $C_3$ has larger contribution than other communities for only \textit{Austin (25)}. \revmert{$C_4$ does not provide a significant contribution to any neighborhood.} For Cincinnati case, \textit{Opioid Overdose Occ.} is the most predictive feature for forecasting future opioid overdose in the same location, which means the local component behaves as an autoregressive module unlike the Chicago case. Furthermore, both violent crimes including \textit{Agg. Assaults, Rape, Homicide, Part 2 Minor (e.g. Menacing)} and property crimes including \textit{Burglary/Breaking Ent., Theft, Part 2 Minor (e.g. Fraud)} are significant features from $C_1$. On the other hand, \textit{Theft} and \textit{Part 2 Minor} from $C_2$, and \textit{Theft} and \textit{Burglary} from $C_3$ are predictive features for future opioid overdose in the target locations. Recall that $C_2$ and $C_3$ have more salient contribution on most of the neighborhoods, which implies that commitment of previous property crimes (especially \textit{Theft}) in the members of those communities may be one of the potential indicators of future opioid overdose in the other neighborhoods. \revmert{Our findings are consistent with the literature that highlighted the connection between crime and drug use, and suggested the property crimes such as theft, burglary might be committed to raise funds to purchase drugs \cite{bennett2008statistical}.}

\begin{figure*}[t!]
\centering
\includegraphics[clip, trim=0cm 2.2cm 3cm 2.9cm, width=.75\textwidth]{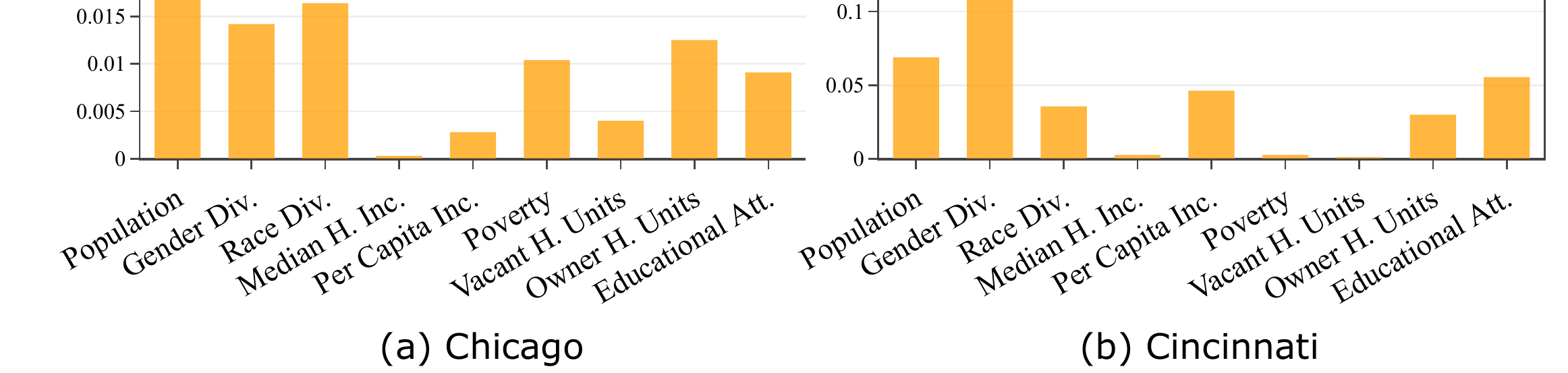}
\caption{\label{fig:05_static_features}
\textbf{Importance of static features.} Mean absolute values of input weights of FC layer in static component.}
\end{figure*}

We explore the importance of \textbf{static features} by analyzing mean absolute input weights of FC in static component (see Fig.~\ref{fig:05_static_features}). For Chicago, demographic features (\textit{Population, Gender Div.} and \textit{Race Div.}) are significant. \revmert{\textit{Owner Occupied H. units, Poverty} and \textit{Educational Att.} are also informative. For Cincinnati, \textit{Gender Div.} and \textit{Population} are important as well as \textit{Educational Att.} and \textit{Per Capita Income}.}  Based on the results, the neighborhoods with higher population, and lower or moderate gender diversity may require additional resources to prevent opioid overdose in both cities. \revmert{Economic status is important for both cities, which is consistent with the previous work suggesting that communities with a higher concentration of economic stressors may be vulnerable to abuse of opioids as a way to manage stress \cite{king2014determinants}. Among three economic status indicators, GL selects only one, \textit{Poverty} for Chicago and \textit{Per Capita Income} for Cincinnati.}
\section{Discussion and Future Work}
\revmert{We presented a community-attentive spatio-temporal model to forecast opioid overdose from crime dynamics. We developed a novel deep architecture based on multi-head attentional networks that learns different representation subspaces and allows the target locations to select location-specific community contributions for forecasting local incidents. Meanwhile, it allows for interpreting predictive features in both local-level and community-level, as well as community memberships and community contributions. We showed the strength of our method through extensive experiments. Our method achieved superior forecasting performance on two real-world opioid overdose datasets compared to baselines.}

\revmert{Our results suggest different spatio-temporal crime-overdose potential links. The overdose deaths at a target neighborhood in Chicago appear to be better predicted by crime incidents at neighborhoods in the same community. Also, change in crime incidences in neighborhoods with low crime rates is an important indicator of future overdoses in most of the other neighborhoods. In Cincinnati, the crime incidents occurred in communities comprising those crime hot-spots seem to well predict the overdose events in most of the neighborhoods. Furthermore, the predictive local activities are different in two cases. While the local crime incidents, \textit{Narcotics} and \textit{Assault}, are predictive for local overdose deaths in Chicago, previous overdose occurrences are informative for future overdose incidents in Cincinnati. On the other hand, the global contributions to forecasting local overdose incidents show similar patterns in both cities. Change in property crimes, in particular \textit{Theft, Deceptive Practice, Burglary} and \textit{Weapons Violation} (crime against to society) in Chicago, \textit{Theft} and \textit{Burglary} in Cincinnati, can be significant indicators for future local overdose incidents as well as certain type of violent crimes (\textit{Battery} for Chicago and \textit{Agg. Assault for Cincinnati}). Last but not the least, demographic characteristics, economic status and educational attainment of the neighborhoods in both cities may help forecasting future local incidents. Our findings support the hypothesis that criminal activities and opioid overdose incidents may reveal spatio-temporal lag effects, and are consistent with the literature. As future work, we plan to investigate the link between opioid use and other social phenomena using our method. We also plan to extend our model to consider multi-resolution spatio-temporal dynamics for prediction.}

\noindent{\bf Acknowledgement.}
This work is part of the research associated with NSF \#1637067 and \#1739413. Any opinions, findings, and conclusions or recommendations expressed in this material do not necessarily reflect the views of the funding sources.

%
%
%

\begin{thebibliography}{8}
\bibitem{bennett2008statistical}
Bennett, T., Holloway, K., Farrington, D.: The statistical association between drug misuse and crime: A meta-analysis. Aggression and Violent Behavior \textbf{13}(2),  107--118 (2008)

\bibitem{burke2016forecasting}
Burke, D.S.: Forecasting the opioid epidemic. Science (354), ~529 (2016)

\bibitem{ertugrul2019}
Ertugrul, A.M., Lin, Y.R., Chung, W.T., Yan, M., Li, A.: Activism via attention: interpretable spatiotemporal learning to forecast protest activities. EPJ Data Science  \textbf{8}(1), ~5 (2019)

\bibitem{ertugrul2018forecasting}
Ertugrul, A.M., Lin, Y.R., Mair, C., Taskaya~Temizel, T.: Forecasting heroin overdose occurrences from crime incidents. In: SBP-BRiMS (2018)

\bibitem{ghaderi2017deep}
Ghaderi, A., Sanandaji, B.M., Ghaderi, F.: Deep forecast: Deep learning-based spatio-temporal forecasting. arXiv preprint arXiv:1707.08110  (2017)

\bibitem{gruenewald2013b}
Gruenewald, P.J.: Geospatial analyses of alcohol and drug problems: empirical needs and theoretical foundations. GeoJournal \textbf{78}(3),  443--450 (2013)

\bibitem{hammersley1989relationship}
Hammersley, R., Forsyth, A., Morrison, V., Davies, J.B.: The relationship between crime and opioid use. Addiction  \textbf{84}(9),  1029--1043 (1989)

\bibitem{hochreiter1997}
Hochreiter, S., Schmidhuber, J.: Long short-term memory. Neural Comput. \textbf{9}(8),  1735--1780 (1997)

\bibitem{huang2018deepcrime}
Huang, C., Zhang, J., Zheng, Y., Chawla, N.V.: Deepcrime: Attentive hierarchical recurrent networks for crime prediction. In: ACM CIKM. pp. 1423--1432 (2018)

\bibitem{jalal2018changing}
Jalal, H., Buchanich, J.M., Roberts, M.S., Balmert, L.C., Zhang, K., Burke, D.S.: Changing dynamics of the drug overdose epidemic in the united states from 1979 through 2016. Science  \textbf{361}(6408), eaau1184 (2018)

\bibitem{kennedy2016opioid}
Kennedy-Hendricks, A., Richey, M., McGinty, E.E., Stuart, E.A., Barry, C.L., Webster, D.W.: Opioid overdose deaths and florida’s crackdown on pill mills. American journal of public health  \textbf{106}(2),  291--297 (2016)

\bibitem{king2014determinants}
King, N.B., Fraser, V., Boikos, C., Richardson, R., Harper, S.: Determinants of increased opioid-related mortality in the united states and canada, 1990--2013: a systematic review. American journal of public health \textbf{104}(8),  e32--e42 (2014)

\bibitem{kolodny2015prescription}
Kolodny, A., Courtwright, D.T., Hwang, C.S., Kreiner, P., Eadie, J.L., Clark, T.W., Alexander, G.C.: The prescription opioid and heroin crisis: a public health approach to an epidemic of addiction. Annual review of public health \textbf{36},  559--574 (2015)

\bibitem{liang2018geoman}
Liang, Y., Ke, S., Zhang, J., Yi, X., Zheng, Y.: Geoman: Multi-level attention networks for geo-sensory time series prediction. In: IJCAI. pp. 3428--3434 (2018)

\bibitem{pierce2015quantifying}
Pierce, M., Hayhurst, K., Bird, S.M., Hickman, M., Seddon, T., Dunn, G., Millar, T.: Quantifying crime associated with drug use among a large cohort of sanctioned offenders in england and wales. Drug \& Alcohol Dependence \textbf{155},  52--59 (2015)

\bibitem{qin2017dual}
Qin, Y., Song, D., Cheng, H., Cheng, W., Jiang, G., Cottrell, G.W.: A dual-stage attention-based recurrent neural network for time series prediction. In: AAAI. pp. 2627--2633 (2017)

\bibitem{rudd2016increases}
Rudd, R.A., Aleshire, N., Zibbell, J.E., Matthew~Gladden, R.: Increases in drug and opioid overdose deaths—united states, 2000--2014. American Journal of Transplantation \textbf{16}(4), 1323--1327 (2016)

\bibitem{scardapane2017group}
Scardapane, S., Comminiello, D., Hussain, A., Uncini, A.: Group sparse regularization for deep neural networks. Neurocomputing \textbf{241}, 81--89 (2017)

\bibitem{seddon2005drugs}
Seddon, T.: Drugs, crime and social exclusion: social context and social theory in british drugs--crime research. British Journal of Criminology \textbf{46}(4),  680--703 (2005)

\bibitem{vaswani2017}
Vaswani, A., Shazeer, N., Parmar, N., Uszkoreit, J., Jones, L., Gomez, A.N., Kaiser, {\L}., Polosukhin, I.: Attention is all you need. In: NIPS. pp. 5998--6008 (2017)

\bibitem{warner2011drug}
Warner, M., Chen, L.H., Makuc, D.M., Anderson, R.N., Mini{\~n}o, A.M.: Drug poisoning deaths in the united states, 1980-2008. NCHS data brief (81), ~1--8 (2011)

\bibitem{zhao2018distant}
Zhao, L., Wang, J., Guo, X.: Distant-supervision of heterogeneous multitask learning for social event forecasting with multilingual indicators. In: AAAI (2018)

\end{thebibliography}
%

\end{document}